\documentclass[journal]{IEEEtran}

\usepackage[numbers,sort&compress]{natbib}

\usepackage{subfigure} 
\usepackage{amsmath,amssymb,amsfonts}
\usepackage{algorithmic}
\usepackage{graphicx}
\usepackage{textcomp}
\usepackage{xcolor}
\usepackage{multirow}
\usepackage{algorithm}
\usepackage{algorithmic}

\ifCLASSINFOpdf
\else
\fi
\begin{document}
%

\title{Double-Flow-based Steganography without Embedding for Image-to-Image Hiding}

%

\author{Bingbing Song, Derui Wang,~\IEEEmembership{Member,~IEEE,} Tianwei Zhang,~\IEEEmembership{Member,~IEEE,} Renyang Liu, Yu Lin* and Wei Zhou* ~\IEEEmembership{Member,~IEEE,}

\thanks{Bingbing Song and Renyang Liu are with the School of Information Science $ \& $ Engineering, Yunnan University, Yunnan, China.}
\thanks{Derui Wang is with Swinburne University of Technology, Melbourne, Australia.}
\thanks{Tianwei Zhang is with School of Computer Science and Engineering, Nanyang Technological University, Singapore.}
\thanks{Yu Lin is with Kunming Institute of Physics, Yunnan, China.}
\thanks{Wei Zhou is with the National Pilot School of Software and Engineering Research Center of Cyberspace, Yunnan University, Yunnan, China.}


}

\maketitle

\begin{abstract}
As an emerging concept, steganography without embedding (SWE) hides a secret message without directly embedding it into a cover. Thus, SWE has the unique advantage of being immune to typical steganalysis methods and can better protect the secret message from being exposed.
However, existing SWE methods are generally criticized for their poor payload capacity and low fidelity of recovered secret messages.
In this paper, we propose a novel steganography-without-embedding technique, named DF-SWE, which addresses the aforementioned drawbacks and produces diverse and natural stego images.
Specifically, DF-SWE employs a reversible circulation of double flow to build a reversible bijective transformation between the secret image and the generated stego image. Hence, it provides a way to directly generate stego images from secret images without a cover image.
Besides leveraging the invertible property, DF-SWE can invert a secret image from a generated stego image in a nearly lossless manner and increases the fidelity of extracted secret images.
To the best of our knowledge, DF-SWE is the first SWE method that can hide large images and multiple images into one image with the same size, significantly enhancing the payload capacity.
According to the experimental results, the payload capacity of DF-SWE achieves $24-72 BPP$ is $8000 \sim 16000$ times compared to its competitors while producing diverse images to minimize the exposure risk.
Importantly, DF-SWE can be applied in the steganography of secret images in various domains without requiring training data from the corresponding domains. This domain-agnostic property suggests that DF-SWE can 1) be applied to hiding private data and 2) be deployed in resource-limited systems.

\end{abstract}

\begin{IEEEkeywords}
Image steganography, Steganography without embedding, Encryption, Flow-based Model, Security.
\end{IEEEkeywords}

%
\IEEEpeerreviewmaketitle

\section{Introduction}
\label{section:1}


\IEEEPARstart{D}{eep} Image steganography aims to concealing secret messages into cover images imperceptibly. The secret messages are only allowed to be recovered by the informed receiver while being invisible to others, which secures its transmission without being noticed \cite{DBLP:conf/eccv/ZhuKJF18,DBLP:journals/pami/Baluja20}. Henceforth, image steganography has been applied in various domains, such as information security \cite{DBLP:journals/corr/abs-1712-06951}, data communication \cite{DBLP:conf/eccv/ZhuKJF18} and copyright protection \cite{RGvanSchyndel1994ADW}.

In the image steganography task, the primary requirements converge to capacity, extraction error, and security.
The embedded steganography (ES) generally select an existing image as a cover and then embed secret information into the cover image with a slight modification.
However, these traditional ES steganography methods \cite{RGvanSchyndel1994ADW,DBLP:conf/issre/WangZJ21} have limited payload capacity.
To further increase payload capacity, deep learning-based ES steganography methods have been proposed recently to achieve both acceptable imperceptibility and small extraction error of secret message \cite{DBLP:conf/nips/Baluja17}. 
However, since all these ES methods need to modify the cover image, the modified cover image always contains a subtle pseudo-shadow of the secret message, especially under a high hiding payload. 
This leads to potential risks of exposing the secret message through compromising the cover image using steganalysis tools.

Instead of directly embedding the secret message into a cover image, steganography without embedding (SWE) is an emerging concept of hiding a secret message without a cover image, which eliminates the modification traces observed in ES methods. 
Thus, SWE has the unique advantage of reducing the risk of secret messages breach from typical steganalysis \cite{DBLP:conf/icccsec/ZhouSHCS15}.
Although current SWE approaches have achieved remarkable results, there still exist some fatal drawbacks.
There are two types of SWE techniques. 1) Mapping-based methods transform the secret message into a sequence of image hashes selected from an existing image set \cite{2016Coverless,DBLP:conf/icic/ZhengWLH17}.
These mapping-based methods require the construction of fixed image mapping rules, which do not accommodate the dynamic growth of images.
2) Alternatively, generating-based methods synthesize images by passing the secret message into a deep generator network, e.g., generative adversarial network (GAN) \cite{ZhuoZhang2018Generative,DBLP:journals/ppna/YuHZJLZ21}.
However, due to the instability of the generative network and the irreversibility of the generative process, a critical weakness is that the payload capacity is extremely limited, especially for hiding large secret images.
As shown in Table \ref{Statistics}, the maximum hiding capacity of the existing works without embedding is 4, and the hiding type can only be bit. In order to realize image-to-image steganography without embedding, the hiding capacity must be at least 24 BPP. If for multi-image hiding, it needs a higher hiding capacity.
Moreover, it is difficult to minimize the message extraction error while keeping the visual quality of the generated stego images \cite{DBLP:journals/ppna/YuHZJLZ21}.

\begin{table}[ht]  
\scriptsize
\renewcommand\arraystretch{1.2}
\caption{The Statistics of Hidden Capacity.}
\begin{center}
\begin{tabular}{c|c|c|c}
\hline
{Methods} & {Year} &{Hiding Type} &{Max payloads (BPP)}\\
\hline
DCGAN-Steg \cite{DBLP:journals/access/HuWJZL18}& 2018 & bit & 9.1e-3 \\
SSteGAN \cite{DBLP:conf/iconip/WangGWQL18}& 2018 & bit  & 2.9e-1\\
SAGAN-Steg \cite{DBLP:journals/ppna/YuHZJLZ21}& 2021 & bit & 4e-1 \\
IDEAS \cite{DBLP:conf/cvpr/0001M0ZS022}& 2022 & bit & 2.3e-2 \\
S2IRT \cite{DBLP:journals/corr/abs-2203-06598}& 2022 & bit & 4 \\
\hline
\end{tabular}
\end{center}
\label{Statistics}
\end{table}

In this paper, we propose a novel double-flow-based steganography without embedding (DF-SWE) approach to tackle the above issues of current SWE methods.
DF-SWE builds a reversible bijective transformation between the secret images and the generated stego images via the invertibility of the flow model and the reversible circulation of double flow.
Our approach significantly enhances the payload capacity and can hide large images without cover images. 
Furthermore, DF-SWE can hide multiple secret images at one time, which greatly extends the capability of SWE-based methods.
To the best of our knowledge, this is the first SWE method working towards image data rather than small binary bits.
In addition, DF-SWE reduces the extraction error, which is attributed to the reversibility of the hiding and restoring processes by the invertible bijective mapping.
Furthermore, DF-SWE guarantees the quality of the generated stego images to enhance the imperceptibility of the secret images.

In sum, our novel DF-SWE method achieves state-of-the-art steganographic performance in the payload capacity, extraction error, and stealthiness of hiding large images. Intriguingly, DF-SWE shows a capability of domain generalization, which makes it applicable to privacy-critical, resource-limited scenarios. 
The detailed contributions are as follows:
\begin{itemize}
    \item \textbf{High payload capacity:} 
    DF-SWE works towards image-to-image generative steganography.
    Our payload capacity (BPP) achieves $24-72 BPP$ and is $8000 \sim 16000 $ times more than existing SWE methods.
    Moreover, DF-SWE is the first method achieving multiple secret images hiding without embedding.
    
    \item \textbf{Low extraction error:} 
    We propose the reversible circulation of double flow to build a reversible bijective transformation between secret images and generated stego images. 
    It is worth noting that reversible circulation of double flow is an invertible process. Hence, we can invert a secret image from a stego image in a nearly lossless manner.
  
    \item \textbf{Enhanced stealthiness:} According to the experimental results, the proposed DF-SWE shows better hiding performance, providing diverse and realistic images to minimize the exposure risk compared to the prior steganography works. Meanwhile, our proposed SWE also achieves better security performance against steganalysis detections.
    
    \item \textbf{Domain generalization:} Our experiments show that, once trained, DF-SWE can be applied in the steganography of secret images from different domains without further model training or fine tuning. This property makes DF-SWE the first domain-agnostic steganography method which can be applied to unseen private data and be executed on resource-limited systems.
\end{itemize}

This paper is organized as follows. Section \ref{Related Work} introduces the related work. Section \ref{Backbone network} briefly describes the Glow model as a backbone network. Section \ref{Methodology} elaborates the proposed DF-SWE method. Section \ref{Experiment} presents and discusses the experimental results. Discussion and future work are drawn in Section \ref{Discussion and Future Work}.

\section{Related Work}
\label{Related Work}

Most existing steganographic approaches are embedded steganography (ES), which embeds the secret information imperceptibly into a cover image by slightly modifying its content. However, the modification traces of the embedded steganography will cause some distortion in the stego image, especially when embedding color image data that usually contain thousands of bits, making them easily detected by steganalysis.
Steganography without embedding is proposed to improve security, which doesn't need to modify the cover image.

\subsection{Embedded steganography}

\textbf{Traditional ES methods:} The Least Significant Bits (LSB) \cite{DBLP:journals/pr/ChanC04} only modified the information of the last few bits, so it would not cause visible changes in the pixel values of the picture. 
In addition, LSB also had many variations \cite{DBLP:journals/spl/Mielikainen06a,DBLP:conf/iciot3/ElHarroussAA20a}. For example, an information hiding technique \cite{DBLP:journals/mta/SahuG22} has been proposed by utilizing the least significant bits (LSBs) of each pixel of grayscale image adopting XOR features of the host image pixels.
Besides, HUGO \cite{DBLP:conf/ih/PevnyFB10} was proposed, and the main design principle was to minimize the properly defined distortion through an efficient coding algorithm.
There are steganographic algorithms not only in the spatial domain, but also in the frequency domain, such as J-UNIWARD \cite{DBLP:journals/ejisec/HolubFD14}, UED \cite{DBLP:journals/tifs/GuoNS14}, 
I-UED\cite{DBLP:conf/icccsec/PanNS16},
UERD \cite{DBLP:journals/tifs/GuoNSTS15}.

\textbf{Deep learning-based ES methods:}
Baluja \cite{DBLP:conf/nips/Baluja17} proposed an autoencoder architecture placing a full-size image in another image of the same size.
After this, Wu et al. \cite{DBLP:conf/pcm/WuYL18} proposed an encoder-decoder architecture, where the cover image and the secret image were concatenated using Separable Convolution (SCR) with residual block.
Besides, Zhang et al. \cite{DBLP:conf/ih/ZhangZCLLY18} combined the method of adversarial examples for steganography.
Replacing encoder-decoder architecture. CycleGAN-based methods \cite{DBLP:journals/scpe/KuppusamyRRSD20,DBLP:conf/cvpr/PoravM019} had proposed for image steganography. 
Furthermore, Zhang et al. \cite{DBLP:journals/mta/ZhangDL19} proposed ISGAN, which improved the invisibility by hiding the secret image only in the Y channel of the cover image.
Wang et al. \cite{DBLP:conf/issre/WangZJ21} designed a multi-level feature fusion procedure based on GAN to capture texture information and semantic features. 
Recently, Invertible Network was proposed for image hiding. Due to the reversible nature of Invertible Network, HiNet \cite{DBLP:conf/iccv/Jing0XWG21}
significantly improves the restored quality of secret image. Based on this, DeepMIH \cite{DBLP:journals/pami/GuanJ0XJZL23} was proposed to hide Multiple Image and achieved excellent performance compared with ES methods.

\subsection{Steganography without embedding (SWE)}

\textbf{Mapping-based SWE methods:}
In 2016, a bag-of-words (BOW) model was proposed to construct the mapping relationship between the dictionary and the words~\cite{2016Coverless}.
Furthermore, Zheng et al. \cite{DBLP:conf/icic/ZhengWLH17} proposed robust image hashing, which calculated the scale-invariant feature transform (SIFT) points in 9 sub-images respectively.
Cao et al. \cite{c53} divided the pixel values from 0 to 255 into 16 intervals, and built a mapping relationship with the bit string of length 4. 
After this, Qiu et al. \cite{2019Coverless} first hashed the local binary pattern (LBP) features of the cover image and the secret image, and then the hashes were matched to create the hidden image.
Besides, a CIS algorithm based on DenseNet feature mapping was proposed \cite{DBLP:journals/ejivp/LiuXQTQ20}, which introduced deep learning to extract high-dimensional CNN features mapped into hash sequences.
Based on GAN, a Star Generative Adversarial Network (StarGAN) was proposed to construct a high-quality stego image with the mapping relationship \cite{DBLP:journals/tnse/ChenZQXX22}.

\textbf{Generating-based SWE methods:}
Stego-ACGAN was proposed to generate new meaning normal images for hiding and extracting information \cite{ZhuoZhang2018Generative}.
In 2018, Hu et al. \cite{DBLP:journals/access/HuWJZL18} mapped secret information into noise vectors and used DCGAN to generate stego image.
After this, Zhu et al. \cite{c63} proposed a coverless image steganography method based on the orthogonal generative adversarial network, adding constraints to the objective function to make the model training more stable. 
For improving the steganography capacity and image quality, A GAN steganography without embedding that combines adversarial training techniques was proposed \cite{DBLP:conf/acmturc/JiangHY0Z20}.
And then, the attention-GAN model was proposed for steganography without embedding
\cite{DBLP:journals/ppna/YuHZJLZ21}.
Besides, Liu et al. \cite{liu2022image} proposed IDEAS based on GAN, which disentangled an image into two representations for structure and texture and utilized structure representation to improve secret message extraction.
Different from GAN-based approachs, Generative Steganographic Flow (GSF)
\cite{DBLP:conf/icmcs/Wei0SZQ022} built a reversible bijective mapping between the input secret data and the generated stego images and took the stego image generation and secret data recovery process as an invertible transformation.
After this, Zhou et al. \cite{DBLP:journals/corr/abs-2203-06598} proposed a secret-to-image reversible transformation (S2IRT), where a large number of elements of given secret message were arranged into the corresponding positions to construct a high-dimensional vector. And then, the vector is mapped to a generated image.

\subsection{Comparison with DF-SWE}

Unlike those SWE method, we propose DF-SWE to hide the secret images rather than the limited binary bits, bringing higher hiding capacity without losing the naturalness of the stego images. 
Meanwhile, we build a reversible bijective transformation between the secret images and the generated stego images, reducing the extraction error of the secret images.

\section{Backbone network}
\label{Backbone network}

In this paper, we propose a double-flow-based model to build a reversible bijective transformation between secret images and generated stego image.
Our flow-based backbone network relies on Glow \cite{DBLP:conf/nips/KingmaD18}.
The flow-based model is commonly used in an image generation tasks by learning a bijective mapping between the latent space of simple distributions and the image space with complex distributions.

In flow-based generative models, the generative process is defined as:
\begin{equation}
    \textsl{z} \sim \textsl{p}_{\theta}(\textsl{z}),
\end{equation}
\begin{equation}
    \textsl{x} \sim  \textsl{g}_{\theta}(\textsl{z}),
    \label{eq2}
\end{equation}
where \textsl{z} is the latent variable and $\textsl{p}_{\theta}(\textsl{z}) $ is usually a multivariate Gaussian distribution $\textsl{N}(\textsl{z};0,I)$.
The function $ \textsl{g}_{\theta}(\cdot) $ is invertible, such that given a datapoint \textsl{x}, latent-variable inference is done by $ \textsl{z}= \textsl{f}_{\theta}(\textsl{x}) =  \textsl{g}_{\theta}^{-1}(\textsl{x}) $.
For brevity, we will omit subscript $\theta$ from $\textsl{f}_{\theta}$ and $ \textsl{g}_{\theta}$.
The function \textsl{f} is composed of a sequence of transformations: $f = f_1 \circ f_2 \circ \dots \circ f_K$ , such that the relationship between x and z can be written as:
\begin{equation}
\textsl{x} \stackrel{f_1}{\longleftrightarrow} h_1 \stackrel{f_2}{\longleftrightarrow} h_2 \dots \stackrel{f_k}{\longleftrightarrow}  \textsl{z},
\end{equation}
where $f_{i} $ is a reversible transformation function and $h_{i}$ is the output of $f_{i} $.

Under the change of variables of Equation~\ref{eq2}, the probability density function of the model for a given a datapoint can be written as
\begin{equation}
\log\textsl{p}_{\theta}(\textsl{x}) = \log\textsl{p}_{\theta}(\textsl{z}) + \log|det(\textsl{dz}/\textsl{dx}).
\end{equation}

The network architectures of Glow comprises three modules, namely the squeeze module, the flow module, and the split module. 
The squeeze module is used to downsample the feature maps, and the flow module is used for feature processing. The split module will divide the image features into halves along the channel side, and half of them are outputted as the latent tensor.

\section{Methodology}
\label{Methodology}

DF-SWE builds a reversible circulation of double flow to generate stego images and hide secret images.
In the reversible circulation of double flow, there are three strategies, \textit{i.e.,} prior knowledge sampling,
high-dimensional space replacement, and distribution consistency transformation.
In the following section, we propose problem definition and threat model. Based on this, we explain the reversible circulation of double flow and the hiding and restoring processes of DF-SWE in detail.

\subsection{Problem definition and threat model}

Given a set $I_{se}:=\{I_{se}\}^k$ of $k$ secret images, a SWE encoder $f_{se}(\cdot):I_{se} \rightarrow z_{se}$ transforms the secret images into random noises $z_{se}$, and $\textsl{z}_{se} \stackrel{t}{\longleftrightarrow}  \textsl{z}_{st}^{\prime}$ is a transformation from $\textsl{z}_{se}$ to $\textsl{z}^{\prime}_{st}$ for secret image hiding. 
In closing, a generator $f_{st}(\cdot):z_{st}\rightarrow\Tilde{I_{se}}$ produces a stego image $\Tilde{I_{se}}$ from the noise $\textsl{z}^{\prime}_{st}$. To maximize the reconstruction performance of the secret images, we propose using an invertible function for both $f_{se}(\cdot)$ and $f_{st}(\cdot)$. 
That is, after taking the inverse $f_{st}^{-1}(\cdot):\Tilde{I_{st}}\rightarrow z_{se}$ and a transformation $\textsl{z}_{st} \stackrel{t^{-1}}{\longleftrightarrow}  \textsl{z}_{se}^{\prime}$  from $\textsl{z}_{st}$ to $\textsl{z}^{\prime}_{se}$,
the secret images can be revealed through $f_{se}^{-1}(\cdot):z_{st}\rightarrow I_{se}$.

In our threat model, the attacker has access to a public training dataset for training the steganography model. During the attacking phase, an attacker gathers the secret images and generates the stego image by a composition $f(\cdot):= f_{se}\circ f_{st}(\cdot)$ of $f_{se}$ and $f_{st}$. Once the stego image is delivered to the recipient, the recipient recovers the secret images by the inverse of the same stego model $f^{-1}(\cdot)$. Moreover, the trained $f(\cdot)$ can be reused for various secret images, even those coming from different domains.

Inspired by Glow \cite{DBLP:conf/nips/KingmaD18}, DF-SWE uses the double-flow-based model to build a reversible bijective transformation between secret images and generated stego images.
The DF-SWE network takes a secret image as its input to generate a realistic stego image.
Later on, it can directly recover the hidden secret image from the stego image via the reversible transformation.

\renewcommand{\dblfloatpagefraction}{.9}
\begin{figure}[ht]
\centering
    \includegraphics[width=.46\textwidth,height=.32\textwidth]{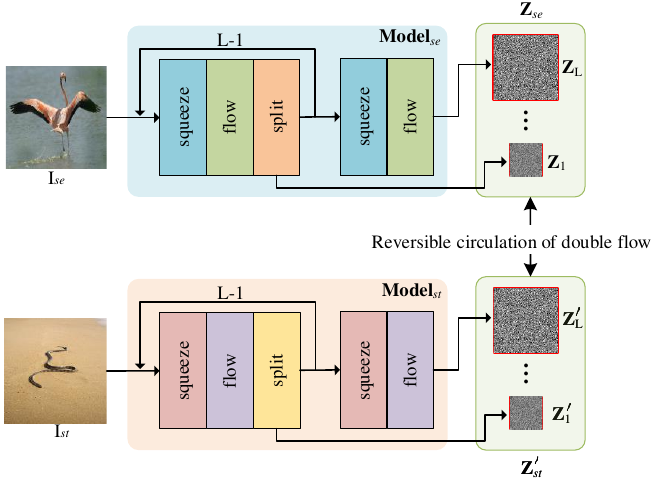}
    \caption{The network architecture of double-flow-based steganography without embedding for image-to-image hiding.}
    \label{figure}
\end{figure}

As illustrated in Figure.~\ref{figure}, the key components of our DF-SWE network are the double-flow-based models and the reversible circulation of double flow. 
A flow-based model ($Model_{se}$) can be regarded as an encoder to encode secret images $I_{se}$ into multivariate Gaussian distributions, while another one ($Model_{st}$) can be seen as a generator to generate stego images $I_{st}$ form multivariate Gaussian distributions. 
Due to the invertibility of the flow model, the two flow models can be considered as decoders to extract secret images.
If we construct a reversible circulation of double flow, $I_{st}$ can be generated by $I_{se}$ and $I_{se}$ can be extracted from $I_{st}$ through our strategy.
More specifically, the latent tensor $ \textsl{z}_{se} = {\textsl{z}_{1}, \textsl{z}_{2}, ..., \textsl{z}_{L}}$ and $ \textsl{z}_{st}^{\prime} = {\textsl{z}_{1}^{\prime}, \textsl{z}_{2}^{\prime}, ..., \textsl{z}_{L}^{\prime}}$. $L$ is the depth of architecture.

We use two Glow models to learn multivariate Gaussian distributions of the secret images $I_{se}$ and the stego image $I_{st}$, separately. 
Given functions $\textsl{f}_{se} := f_1 \circ f_2 \circ \dots \circ f_K$ and $\textsl{f}_{st} := f_1^{\prime} \circ f_2^{\prime} \circ \dots \circ f_n^{\prime}$, we have $ I_{se} \stackrel{f_1}{\longleftrightarrow} h_1 \dots h_{k-1} \stackrel{f_k}{\longleftrightarrow}  \textsl{z}_{se} $, $ I_{st} \stackrel{f_1^{\prime}}{\longleftrightarrow} h_1^{\prime}  \dots h_{n-1}^{\prime} \stackrel{f_n^{\prime}}{\longleftrightarrow}  \textsl{z}_{st}^{\prime} $.

The existing flow model (Glow) implements a mapping relationship between the distribution of \textsl{z} and that of the generated image. 
In contrast, large image steganography without embedding is a generative task from an image to another. 
Hence, the core task of image-to-image steganography without embedding is to construct a mapping between the secret image $I_{se}$ and the stego image $I_{st}$ while ensuring the mapping is reversible to enhance the extraction quality of $I_{se}$.
This task can be formulated as follows:
\begin{equation}
I_{se} \stackrel{f_1}{\longleftrightarrow} h_1 \dots \stackrel{f_k}{\longleftrightarrow}  \textsl{z}_{se} \stackrel{t}{\longleftrightarrow}  \textsl{z}_{st}^{\prime} \stackrel{f_1^{\prime}}{\longleftrightarrow}  \dots h_{n-1}^{\prime} \stackrel{f_n^{\prime}}{\longleftrightarrow} I_{st}.
\end{equation}

$\textsl{z}_{se} \stackrel{t}{\longleftrightarrow}  \textsl{z}_{st}^{\prime}$ is a transformation from a multivariate Gaussian distributions $\textsl{z}_{se}$ to another multivariate Gaussian distributions $\textsl{z}^{\prime}_{se}$. Consequently, the core task is the transformation $t$ to construct a reversible circulation in the double flow model, for hiding the secret image in the generated stego image and keeping it reversible.

\subsection{Reversible circulation of double flow}

For transmitting $\textsl{z}_{se}$ to $\textsl{z}_{st}^{\prime}$ and keeping it reversible, we divide the task of $\textsl{z}_{se} \stackrel{t}{\longleftrightarrow}  \textsl{z}_{st}^{\prime}$ into three tasks that need to be solved.
\begin{itemize}
    \item \textbf{How to initialize $\textsl{z}_{st}^{\prime}$? }

    \item \textbf{How to transmit $\textsl{z}_{se}$ to $\textsl{z}_{st}^{\prime}$?}

    \item \textbf{How to reduce the distortions on generated stego images?}
\end{itemize}

In order to solve the above issues, we propose three techniques named prior knowledge sampling, high-dimensional space replacement, and distribution consistency transformation. We use the latent variables of $\textsl{z}$ and its variants (e.g., $\textsl{z}_{st}^{\prime}, \hat{\textsl{z}}^{st}$) to describe the circulation of two flows at different stages after different operations.

\subsubsection{Prior knowledge sampling (PKS) }
\ 

For initializing $\textsl{z}_{st}^{\prime}$, we utilize the prior knowledge of the generator of Glow.
Firstly, \textsl{z} is sampled from 
$\textsl{N}(0,I)$ and the generated image $I_{ge}$ is generated from a Glow model $Glow_{g_{st}}(\textsl{z})$.
The process can be formulated as:
\begin{equation}
\begin{aligned}
 I_{ge} &= Glow_{g_{st}}(\textsl{z}),\\
 \textsl{z} &\sim \textsl{N}(0,I).
\end{aligned}
\end{equation}
During the generation of $I_{ge}$, $Glow_{g_{st}}$ utilizes prior knowledge of Glow parameters to generate image and the generation is irreversible.
Next, we obtain the initialized $ \textsl{z}_{st}^{\prime} $ by a sequence of invertible transformations, which can be formulated as:
\begin{equation}
I_{ge} \stackrel{f_1^{\prime}}{\longleftrightarrow} h_1^{\prime}  \dots h_{n-1}^{\prime} \stackrel{f_n^{\prime}}{\longleftrightarrow}  \textsl{z}_{st}^{\prime}.
\end{equation}

\subsubsection{High-dimensional space replacement (HDSR)}
\label{High-dimensional space replacement}
\

For transmitting $ \textsl{z}_{se} $ to $ \textsl{z}_{st}^{\prime} $ and reducing the generated stego image distortion, we proposed the high-dimensional space replacement.

In the backbone network(Glow), each of the $L$ layers of feature maps in $Model_{se}$ is divided into halves along the channel dimension to  two sets.
Half of the sets are outputted as the latent tensor $\{z_i\}_{i=1}^{L}$, and the other half of the sets are cycled into the squeeze module. 
Hence, $\textsl{z}_{se} $ contains different levels of information about the image. As shown in Figure.~\ref{figure}, $\textsl{z}_{se} = \left\{ \textsl{z}_1, \dots ,\textsl{z}_{L-1}, \textsl{z}_{L} \right\}$ and $\textsl{z}_{st}^{\prime} = \left\{ \textsl{z}_1^{\prime}, \dots ,\textsl{z}_{L-1}^{\prime}, \textsl{z}_{L}^{\prime} \right\}$.
Particularly, we find that the latent tensor from shallow layers of $Model_{se}$ has greater effect on the reversibility of the image. 
If $\textsl{z}_{st}^{\prime}$ is replaced with $\textsl{z}_{se}$ directly, it will cause the distortion of the stego image due to the distribution differences between $\textsl{z}_{st}^{\prime}$ and $\textsl{z}_{se}$.

\begin{figure}[ht]
\centering
    \includegraphics[width=.4\textwidth,height=.3\textwidth]{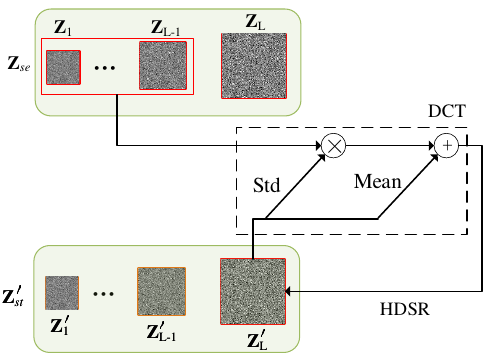}
    \caption{High-dimensional space replacement and Distribution consistency transformation. The Std and Mean are the variance and mean of $\textsl{z}_{L}^{\prime}$. The \textcircled{×} and \textcircled{+} are the matrix
 operations of multiplication and addition respectively. The DCT is the distribution consistency transform and the HDSR is the high-dimensional space replacement.}
    \label{figure2}
\end{figure}

Since different latent tensors of $\textsl{z}_i $ 
have different effects on the reconstruction of the image, we propose high-dimensional space replacement, which replaces the high-dimensional distribution of generated images with the low-dimensional distribution of secret images. 
Our technique follows the principle of minimum information loss. 
As shown in Figure \ref{figure2}, $\textsl{z}_{L}^{\prime} $ is replaced with the concatenated $\left\{ \textsl{z}_1, \dots ,\textsl{z}_{L-1} \right\}$. For brevity, we abbreviate this process as that $ \hat{\textsl{z}}^{st} $ is replaced with $\hat{\textsl{z}}^{se} $. 
The $\textsl{z}_{se} $ of the secret image is circulated to the $\textsl{z}_{st}^{\prime} $ of the stego image, reducing the impact of the secret image and stego image generation. During the secret image extraction phase, $\hat{\textsl{z}}^{se} $ is replaced with $ \hat{\textsl{z}}^{st} $. 

\subsubsection{Distribution consistency transformation (DCT)}
\label{Distribution consistency transformation}
\

High-dimensional space replacement has circulated $ \textsl{z}_{se} $ of secret image to the $ \textsl{z}_{st}^{\prime} $ of the stego image and reduced the generated stego image distortion. 
For further improving the quality of image generation and reducing the generated image distortion, we propose distribution consistency transformation, which can decrease the distribution discrepancy between $\hat{\textsl{z}}^{se} $ and $ \hat{\textsl{z}}^{st} $. 

As shown in Figure \ref{figure2}, distribution consistency transformation is implemented in the high-dimensional space replacement. 
Because flow-based generative models learn reversible bijective transformation between images and a multivariate Gaussian, $\hat{\textsl{z}}^{se} $ and $ \hat{\textsl{z}}^{st} $ obey the Gaussian distribution. Hence, the most important thing to measure the Gaussian distribution is its mean and variance.

Based on this, our proposed distribution consistency transformation is to maintain the consistency of the mean and variance between two distributions. Distribution consistency transformation is defined as follows:

\begin{equation}
Std = \frac{ \sum_{i=1}^{n} \hat{\textsl{z}}^{st}_i- \frac{ \sum_{i=1}^{n}\hat{\textsl{z}}^{st}_i}{n} }{\sum_{i=1}^{n} \hat{\textsl{z}}^{se}_i- \frac{ \sum_{i=1}^{n}\hat{\textsl{z}}^{se}_i}{n}},
\label{equ:std}
\end{equation}

\begin{equation}
Mean = \frac{ \sum_{i=1}^{n} Std \times  \hat{\textsl{z}}^{st}_i - \hat{\textsl{z}}^{se}_i}{n},
\label{equ:mean}
\end{equation}

\begin{equation}
\hat{\textsl{z}}^{se} = \hat{\textsl{z}}^{se} \times Std + Mean .
\label{equ:std+mean}
\end{equation}
Equations (\ref{equ:std}), (\ref{equ:mean}), (\ref{equ:std+mean}) can achieve the reduction of the distribution discrepancy between $\hat{\textsl{z}}^{se} $ and $ \hat{\textsl{z}}^{st} $.
During the secret image extraction phase, the reversible transformation of distribution consistency transformation is expressed as Equation (\ref{equ:12}):

\begin{equation}
\hat{\textsl{z}}^{st} = \frac{\hat{\textsl{z}}^{st} - Mean}{Std}.
\label{equ:12}
\end{equation}


\subsection{Hiding and restoring processes}

\renewcommand{\dblfloatpagefraction}{.9}
\begin{figure*}[htb]
\centering
    \includegraphics[width=.8\textwidth,height=.41\textwidth]{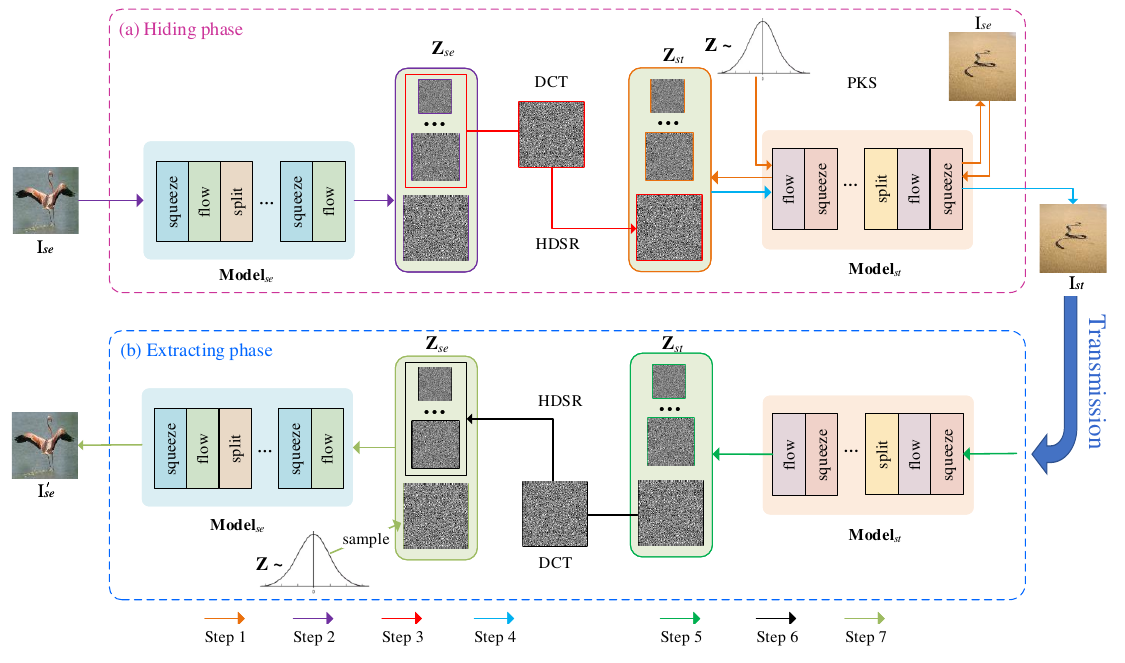}
    \caption{Hiding and restoring processes. (a) is the hiding phase, including Step 1-4. (b) is the extracting phase, including Step 5-7.}
    \label{figure3}
\end{figure*}

In this section, we will describe the secret image hiding and restoring processes in detail. 
As shown in Figure.~\ref{figure3}, DF-SWE comprises two stages, a secret image hiding phase and an extracting phase.

\subsubsection{Hiding process}
Figure.~\ref{figure3} (a) describes the hiding phase, which can hide large images without embedding. 
$ Model_{se} $ and $ Model_{st} $ are two different Glow models.  
Firstly, as shown in Step 1, $ Model_{st} $ randomly samples a Gaussian distribution $\textsl{z} $ to generate an image $I_{ge}$ utilizing prior knowledge of $ Model_{st} $. 
Based on the generated image $I_{ge}$, we use the reversible operation of $ Model_{st} $ to obtain an initialized distribution $ \textsl{z}_{st}^{\prime} $ in order to better carry the secret flow. 
Secondly, as shown in Step 2, $ Model_{se} $ encodes the secret image as $ \textsl{z}_{se} $ by the reversible operation of $ Model_{se} $. Specially, Step 1 and 2 can run in parallel or exchange their sequences.
Through the operation of the high-dimensional space replacement and distribution consistency transformation on Step 3, $ \textsl{z}_{se} $ can be passed to $ \textsl{z}_{st}^{\prime}$ to generate a stego image. Meanwhile, the hiding phase maintains reversibility for extracted secret images.
Finally, $I_{st}$ will be generated by $ \textsl{z}_{st}^{\prime}$ utilizing the $ Model_{st} $ in step 4.

\subsubsection{Restoring process}
As shown in Figure.~\ref{figure3} (b), the extracting phase is the inverse process of hiding phase.
Hence, DF-SWE can extract the secret image with high quality because we construct an invertible mapping of the secret and stego images.
Firstly, the stego image is decoded as $ \textsl{z}_{st}^{\prime}$ by utilizing the reversible operation of $ Model_{st} $ in Step 5. 
And then, through the reverse operation of high-dimensional space replacement and distribution consistency transformation of Step 6, $ \textsl{z}_{st}^{\prime}$ can be passed to $ \textsl{z}_{se} $ to extract the secret image. 
The reverse operation of high-dimensional space replacement and distribution consistency transformation are described in detail in subsection \ref{High-dimensional space replacement} and \ref{Distribution consistency transformation}. Finally, $ Model_{se} $ extracts the secret image $I_{se}^{\prime}$ with high quality in Step 7.

\section{Experimental Results}
\label{Experiment}
\subsection{Experimental setup}
To demonstrate the superiority of DF-SWE, we compare it with four state-of-the-art SWE methods, namely DCGAN-Steg \cite{DBLP:journals/access/HuWJZL18}, SAGAN-Steg \cite{DBLP:journals/ppna/YuHZJLZ21}, SSteGAN \cite{DBLP:conf/iconip/WangGWQL18} and WGAN-Steg \cite{WGAN-GP}. 
To verify the extraction quality of the secret images, we compare DF-SWE with ES methods including 4 bit-LSB, Baluja\cite{DBLP:journals/pami/Baluja20}, Weng et al. \cite{DBLP:conf/mir/WengLCM19} and HiDDeN\cite{DBLP:conf/eccv/ZhuKJF18}, since the existing SWE methods cannot be applied to image data.

Our DF-SWE and baseline models are trained on the datasets of Bedroom (subsets of LSUN, including 3033042 color images) \cite{DBLP:journals/corr/YuZSSX15}, LFW \cite{LFWTechUpdate} (including 13234 color images), and CelebA \cite{liu2015faceattributes} (including 202599 color images).
We train DF-SWE with the hyper-parameter $L= 4$. $L$ is the depth of the model. The greater the depth of the model, the better the quality of the generated images, but the model parameters and computational resources increase. Therefore, the hyper-parameter $L$ can be set according to actual requirements. Besides, the steganography process is completed in less than a second on a GPU RTX3090 with $L= 4$. Therefore, our proposed method has excellent performance in real time applications.

We evaluate the hiding capacity of DF-SWE by comparing the bits per pixel (BPP), BPP = $ \frac{Len(secret)}{H \times W} $, which is the number of message bits hidden per pixel of the encoded image. H/W is the height/width of stego images. Meanwhile, we evaluate the detection error (Pe), $ Pe = \frac{1}{2}(P_{FA}+P_{MD})$. where $P_{FA}$ and $P_{MD}$ represent the probabilities of false alarm and missed detection rate, respectively. $Pe$ ranges in $ [0, 1] $, and its optimal value is 0.5.
As a proxy to secrecy, we can also measure the secret image extraction performance using peak signal-to-noise ratio (PSNR), Root Mean Square Error (RMSE) and Structure Similarity Index Measure (SSIM). A larger value of PSNR, SSIM and smaller value of RMSE indicate higher image quality, which are formulated as follows:

\begin{itemize}
\item \textbf{RMSE:} Root Mean Square Error (RMSE) measures the difference between two images. Given two images $X$ and $Y$ with width $W$ and height $H$, RMSE is formulated as follows:
\begin{equation}
RMSE = \sqrt{MSE},
\label{RMSE}
\end{equation}

\begin{equation}
MSE = \frac{1} {W*H}\sum_{i=1}^{W}\sum_{j=1}^{H}({X_{i,j}-X_{i,j})}^2,
\label{MSE}
\end{equation}
where $X_{i,j}$ and $Y_{i,j}$ indicate the pixels at position $(i,j)$ of images $X$ and $Y$, respectively.

\item \textbf{PSNR:} Peak signal-to-noise ratio (PSNR) is a widely used metric to measure the quality of an image. PSNR is defined as follows:

\begin{equation}
PSNR = 10*\log_{10}{\frac{R^2}{MSE}},
\label{equ:2}
\end{equation}

where $R$ represents the maximum value of images, which is usually set as 255. 

\item \textbf{SSIM: }
Structural Similarity Index Measure (SSIM) is another commonly used image quality assessment based on the degradation of structural information \cite{DBLP:journals/tip/WangBSS04}. SSIM is computed by the means $\mu_{X}$ and $\mu_{X}$, the variance $\sigma_{X}$ and $\sigma_{X}$, and the co-variance $\sigma_{(X,Y)}$, as follows:

\begin{equation}
SSIM = \frac{(2 \mu_{X}\mu_{Y} + C_1) (\sigma_{(X,Y)} + C_2)}{({\mu_{X}}^2+{\mu_{Y}}^2 + C_1) ({\sigma_{X}}^2+{\sigma_{Y}}^2 + C_2)}
\label{SSIM}
\end{equation}

where $c1 = k_{1} L^{2}, c2 = k_{2} L^{2}$ and $L$ is the dynamic range of the pixel values. The default configuration of $k_1$ is 0.01 and $k_2$ is 0.03.

\end{itemize}

\subsection{Evaluation by image hiding quality}

\renewcommand{\dblfloatpagefraction}{.9}
\begin{figure*}[htb]
\centering
    \includegraphics[width=.85\textwidth,height=.47\textwidth]{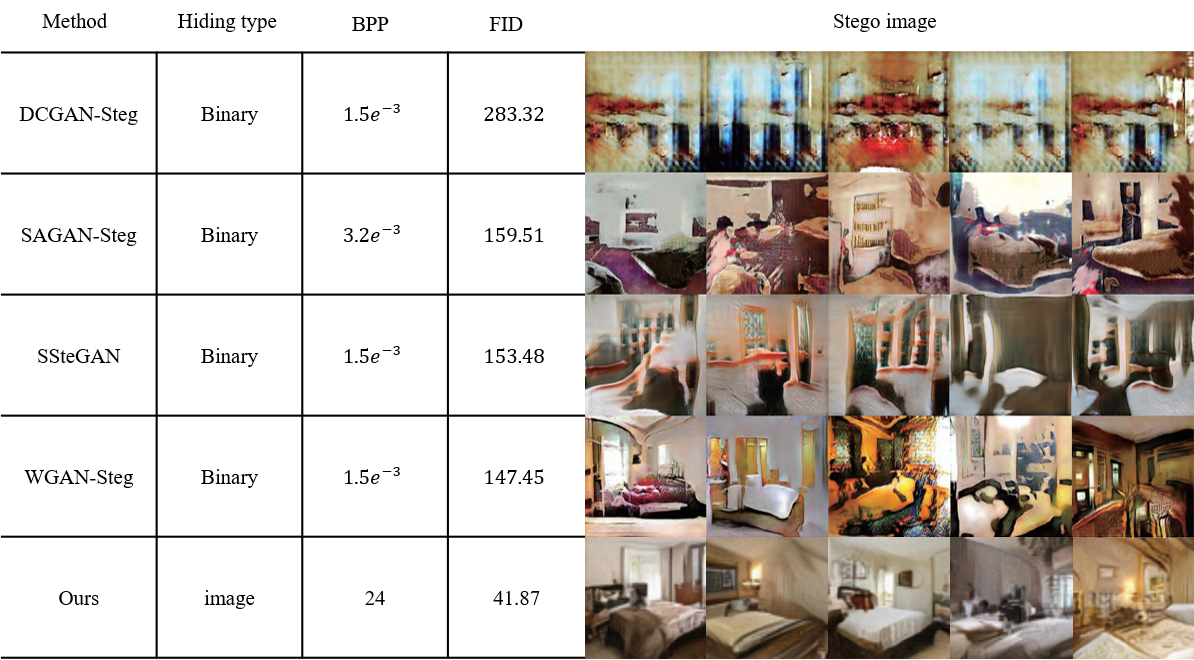}
    \caption{Hiding evaluation with steganography without embedding.}
    \label{fig1}
\end{figure*}

\renewcommand{\dblfloatpagefraction}{.9}
\begin{figure*}[htb]
\centering
    \includegraphics[width=.8\textwidth,height=.42\textwidth]{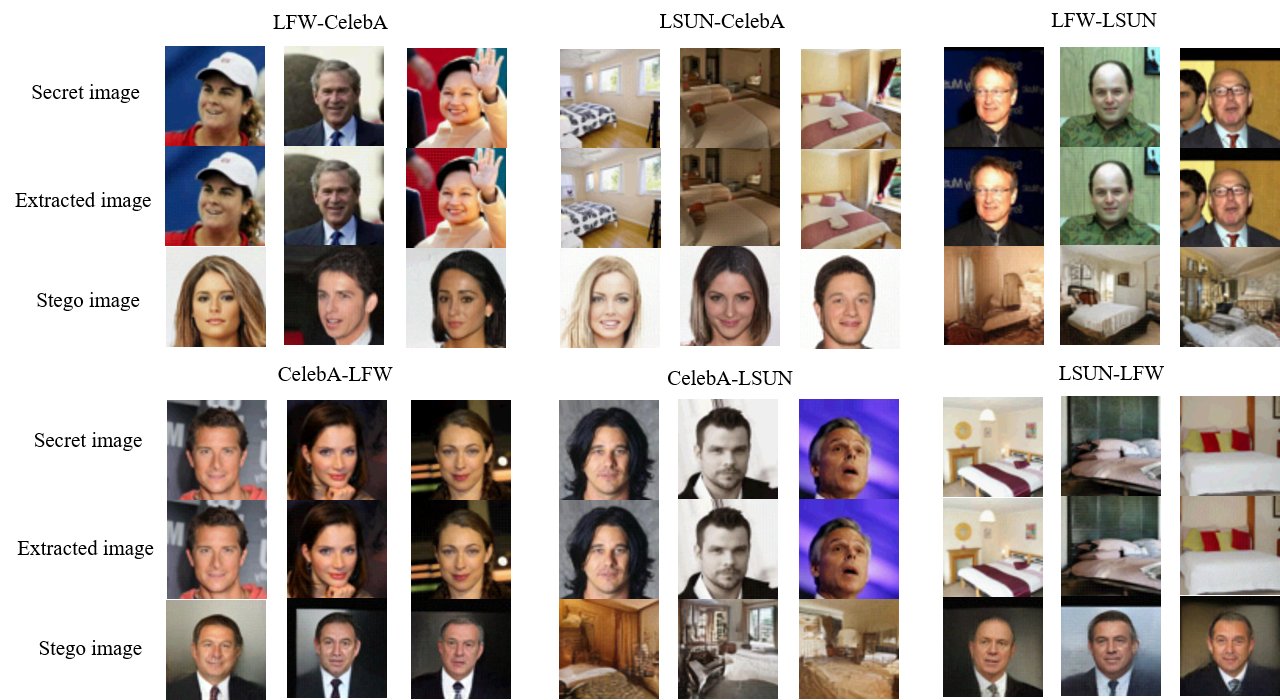}
    \caption{The quality of generated images with the size $ 64 \times 64 $ using DF-SWE.}
    \label{fig2}
\end{figure*}

\renewcommand{\dblfloatpagefraction}{.9}
\begin{figure*}[htb]
\centering
    \includegraphics[width=.6\textwidth,height=.23\textwidth]{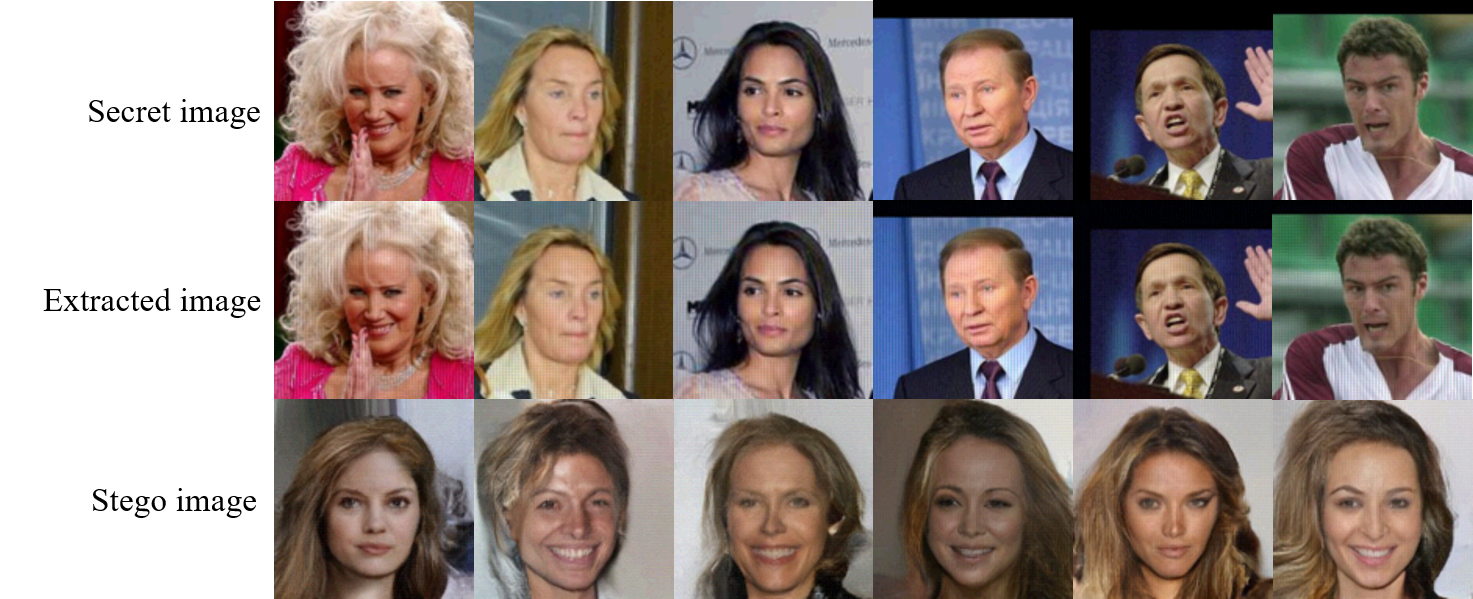}
    \caption{The quality of generated images with the size $ 128 \times 128 $ using DF-SWE.}
    \label{fig3}
\end{figure*}

Figure.~\ref{fig1} is compares our DF-SWE with SWE methods on the bedroom dataset (subsets of LSUN).
Since SWE methods hide secret messages without embedding modifications and are immune to typical steganalysis tools, visual quality is crucial.
From Figure.~\ref{fig1}, we can see that images generated by DF-SWE have higher capacity and are more realistic with the FID (Fréchet Inception Distance) than that of the competitors. FID is a metric of image generation and a lower FID score means that the generated image is more realistic.
There are noticeable distortions in the stego images generated by these SWE methods.
More importantly, in lieu of hiding images, these rivals only support secret messages in binary bits.
The BPP of these SWE methods is around $ 1.5e^{-3} $ while our BPP is $24 $, which is $8000-16000 $ times more than that of the competitors. Hence, our DF-SWE can hide secret images with the same size as the stego images.

Examples of stego images generated by DF-SWE are given in Figure.~\ref{fig2} and Figure.~\ref{fig3}, which show the hiding quality of images in sizes $ 64 \times 64 \times 3 $ and $ 128 \times 128 \times 3 $, respectively. 
It can be observed that the stego images leak no information of the secret images.
Only through $ Modle_{se}, Model_{st} $ and reversible circulation of double flow, the secret image can be extracted from the stego image. 
$ Modle_{se} $ and $ Model_{st} $ have hundreds of millions of parameters and different network structures, which makes decrypting the secret images difficult. 

Once trained, DF-SWE can be generalized to hiding images from various domains.
Figure. \ref{fig2} and Figure. \ref{fig3} show secret images and generated stego images, in different domains. 
For example, the LFW-CelebA signifies that the secret image is randomly selected from the LFW dataset and generated stego image is similar to the style of the CelebA dataset.
From Figure \ref{fig1}, we can see that images generated by DF-SWE are more realistic and extracted secret images have nearly lossless extraction quality.

\subsection{The extraction quality compared with prevalent methods.}

\begin{table*}[ht]  
\scriptsize
\renewcommand\arraystretch{1.2}
\caption{The information extraction accuracy of those methods with different hidden payloads.}
\begin{center}
\begin{tabular}{c|c|cccccc}
\hline
\multirow{2}{*}{Methods} &
\multirow{2}{*}{Type} &
\multicolumn{6}{c}{Hiding payloads (BPP)}\\
\cline{3-8}
& & 1 & 2 & 4 & 6 & 12 & 24 \\
\cline{1-8}
DCGAN-Steg \cite{DBLP:journals/access/HuWJZL18} & bit & 0.7134 & 0.712 & 0.7122 & - & -& - \\
SAGAN-Steg \cite{DBLP:journals/ppna/YuHZJLZ21} & bit & 0.7245 & 0.7232 & 0.723 & - & -& - \\
SSteGAN \cite{DBLP:conf/iconip/WangGWQL18} & bit  & 0.7139 & 0.7126 & 0.7124 & - & -& - \\
WGAN-Steg \cite{WGAN-GP} & bit & 0.7122 & 0.7114 & 0.7113 & - & -& - \\
IDEAS \cite{DBLP:conf/cvpr/0001M0ZS022} & bit & 0.7552 & 0.755 & 0.7546 & - & -& - \\
S2IRT \cite{DBLP:journals/corr/abs-2203-06598} & bit & 1 & 1 & 0.9942 & - & -& - \\
Ours & image &\textbf{1}&\textbf{1}& \textbf{1}&\textbf{0.9921}&\textbf{0.9836} &\textbf{0.5124} \\\cline{1-8}
\end{tabular}
\end{center}
\label{hiding BPP}
\end{table*}

Table \ref{hiding BPP} lists the performances of information extraction accuracy of different steganographic approaches, i.e., DCGAN-Steg, SAGAN-SSteGAN, WGAN-Steg, IDEAS, and S2IRT, with the increase of hiding payloads. From this table, it is clear that DF-SWE achieve much higher information extracted accuracy than SWE approaches under different hiding payloads. The extracted accuracy rates of DF-SWE keep at a very high-level when the hiding payload ranges from 1 BPP to 4 BPP. Besides, the proposed generative steganographic approach can achieve high hiding capacity (up to 12 BPP) and accurate extraction of secret message (almost $100\% $ accuracy rate), simultaneously. 
Even when hiding images (BPP = 24), the extracted accuracy achieve 0.5124 and the pixel errors of the extracted images are mostly ranged in $ \pm 1$. That is because DF-SWE built image-to-image reversible bijective mapping reducing the extraction error. 
In contrast, the extracted accuracy of other SWE methods decrease with the increase of hiding payload.
At high hiding capacity, existing SWE methods cannot hide secret messages or generate stego images are twisted and distorted. Thus, existing SWE methods cannot achieve accurate information extraction under high hiding payloads.

\begin{table*}[ht]  
\scriptsize

\renewcommand\arraystretch{1.2}
\caption{Extraction metrics of secret images compared with prevalent methods.}
\begin{center}
\begin{tabular}{c|c|ccc|ccc|ccc}
\hline
\multirow{2}{*}{Methods} &
\multirow{2}{*}{Type} &
\multicolumn{3}{c|}{LSUN} &\multicolumn{3}{c|}{CelebA} &\multicolumn{3}{c}{LFW}\\
\cline{3-11}
& & PSNR↑ & SSIM↑ &  RMSE↓  & PSNR↑ & SSIM↑ & RMSE↓  & PSNR↑ & SSIM↑ & RMSE↓\\\cline{1-11}
4bit-LSB& ES & 23.06&0.7638& 18.05&23.13&0.7518 &17.86&23.13&0.7668& 17.92\\
Baluja\cite{DBLP:journals/pami/Baluja20} & ES & 32.41&0.9242 &6.31&32.94&0.9325 &6.11&32.61&0.9392 &6.03\\
Weng et al. \cite{DBLP:conf/mir/WengLCM19}& ES  &33.74&0.9518&5.25&34.51&0.9582 &4.98&34.23&0.9657& 4.98\\
HiDDeN\cite{DBLP:conf/eccv/ZhuKJF18}& ES &34.49&0.9536 &4.32&36.34&0.9629 &4.07&36.24&0.9682 &3.96\\
Ours & SWE&\textbf{34.51}&\textbf{0.9542}& \textbf{4.24}&\textbf{37.85}&\textbf{0.9675} &\textbf{2.51}&\textbf{38.13}&\textbf{0.9697} &\textbf{3.21}\\\cline{1-11}
\end{tabular}
\end{center}
\label{table extraction metrics}
\end{table*}

The extraction metrics of the different ES methods are given in Table~\ref{table extraction metrics}, which describes the extraction quality of secret images by PSNR, SSIM and RMSE. 
The columns of LSUN, CelebA and LFW represent the experimental results under different datasets, respectively.
Unlike our DF-SWE to built image-to-image reversible bijective mapping, existing SWE methods directly write the secret message into a latent space and generate the image directly by the latent space. It is difficult to balance the hiding capacity and generation quality.
Since the existing SWE methods face the problem of low hidden capacity and incapability of hiding secret images with the same size, we compared DF-SWE with ES methods to verify the extraction quality of the secret images.
In particular, ES methods usually have a better extraction performance than SWE methods, because ES methods have cover images to hide the secret image and do not consider the generated quality. 
On the contrary, SWE methods require plausible visual quality of both the generated stego image and the recovered secret image. 
From Table~\ref{table extraction metrics}, it is evident that DF-SWE outperforms all other methods, providing better secret image extraction quality.

\subsection{Security evaluation by steganalysis}

We compare our DF-SWE with ES and SWE schemes as shown in Table~\ref{Security evaluation by steganalysis}.
The steganalysis performance is measured by PE metrics.
The optimal value of detection error (Pe) is 0.5. At this time, the steganalyzer (Ye-net~\cite{DBLP:journals/tifs/YeNY17}) cannot distinguish the source of images and can only perform random guess.
Most ES methods have poor steganographic security, while our proposed SWE achieves better security performance with higher Pe values. 
Compared to SWE schemes, DF-SWE has shown significant improvements in several aspects.
The payload is more than 8000 times higher than that of others. 
Meanwhile, we have achieved Pe values better than most of the other works.

\begin{table}[ht]  
\scriptsize
\renewcommand\arraystretch{1.2}
\caption{Security evaluation by steganalysis.}
\begin{center}
\begin{tabular}{c|c|c|c|c}
\hline
\multicolumn{1}{c|}{Type} &
\multicolumn{1}{c|}{Methods} &
\multicolumn{1}{c|}{Hiding type} &\multicolumn{1}{c|}{Payload (BPP) ↑} &\multicolumn{1}{c}{Pe → 0.5}\\
\hline
ES & Baluja\cite{DBLP:journals/pami/Baluja20} &  image & 24 & 0.04\\
ES & Weng et al. \cite{DBLP:conf/mir/WengLCM19}& image  & 24  & 0.04  \\
ES & HiDDeN\cite{DBLP:conf/eccv/ZhuKJF18}&  image  & 24  & 0.03 \\
\hline
SWE & DCGAN-Steg \cite{DBLP:journals/access/HuWJZL18} & Binary & $1.5e^{-3} $  & 0.48 \\
SWE & SAGAN-Steg \cite{DBLP:journals/ppna/YuHZJLZ21} & Binary & $3.2e^{-3}$  & 0.47 \\
SWE & SSteGAN \cite{DBLP:conf/iconip/WangGWQL18} & Binary & $1.5e^{-3}$  & 0.52 \\
SWE & WGAN-Steg \cite{WGAN-GP} & Binary & $1.5e^{-3}$  & 0.52 \\
SWE & Ours &image & 24  & 0.51 \\
\hline
\end{tabular}
\end{center}
\label{Security evaluation by steganalysis}
\end{table}

\subsection{Multiple image hiding}
\renewcommand{\dblfloatpagefraction}{.9}
\begin{figure*}[htb]
\centering
    \includegraphics[width=.9\textwidth,height=.4\textwidth]{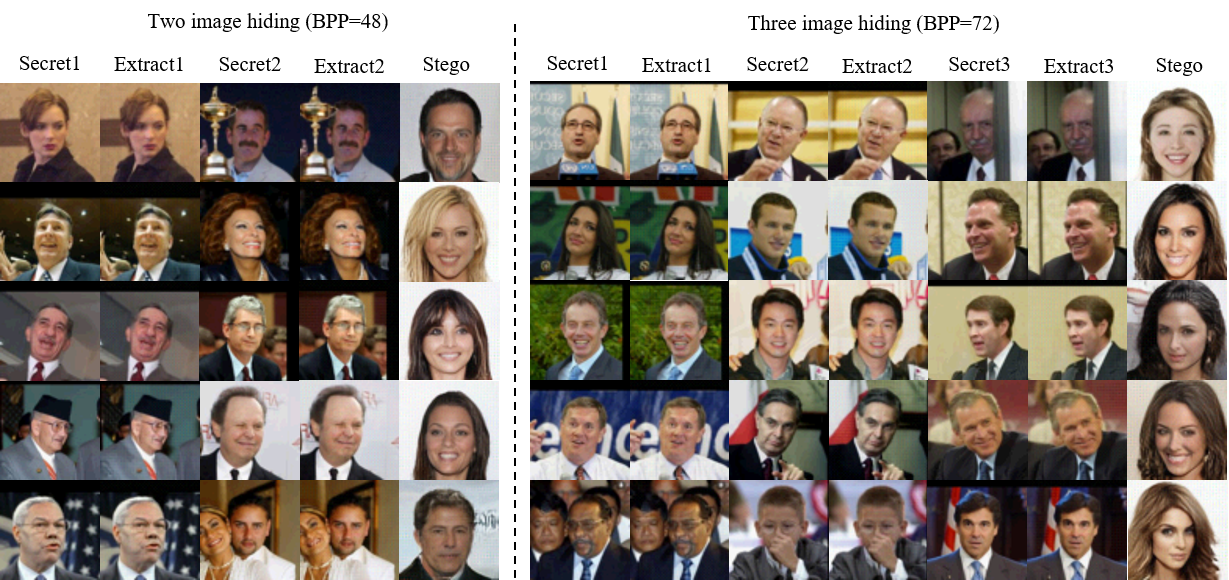}
    \caption{Multiple image hiding.}
    \label{fig5}
\end{figure*}
\renewcommand{\dblfloatpagefraction}{.9}
\begin{figure*}[htb]
\centering
    \includegraphics[width=.6\textwidth,height=.35\textwidth]{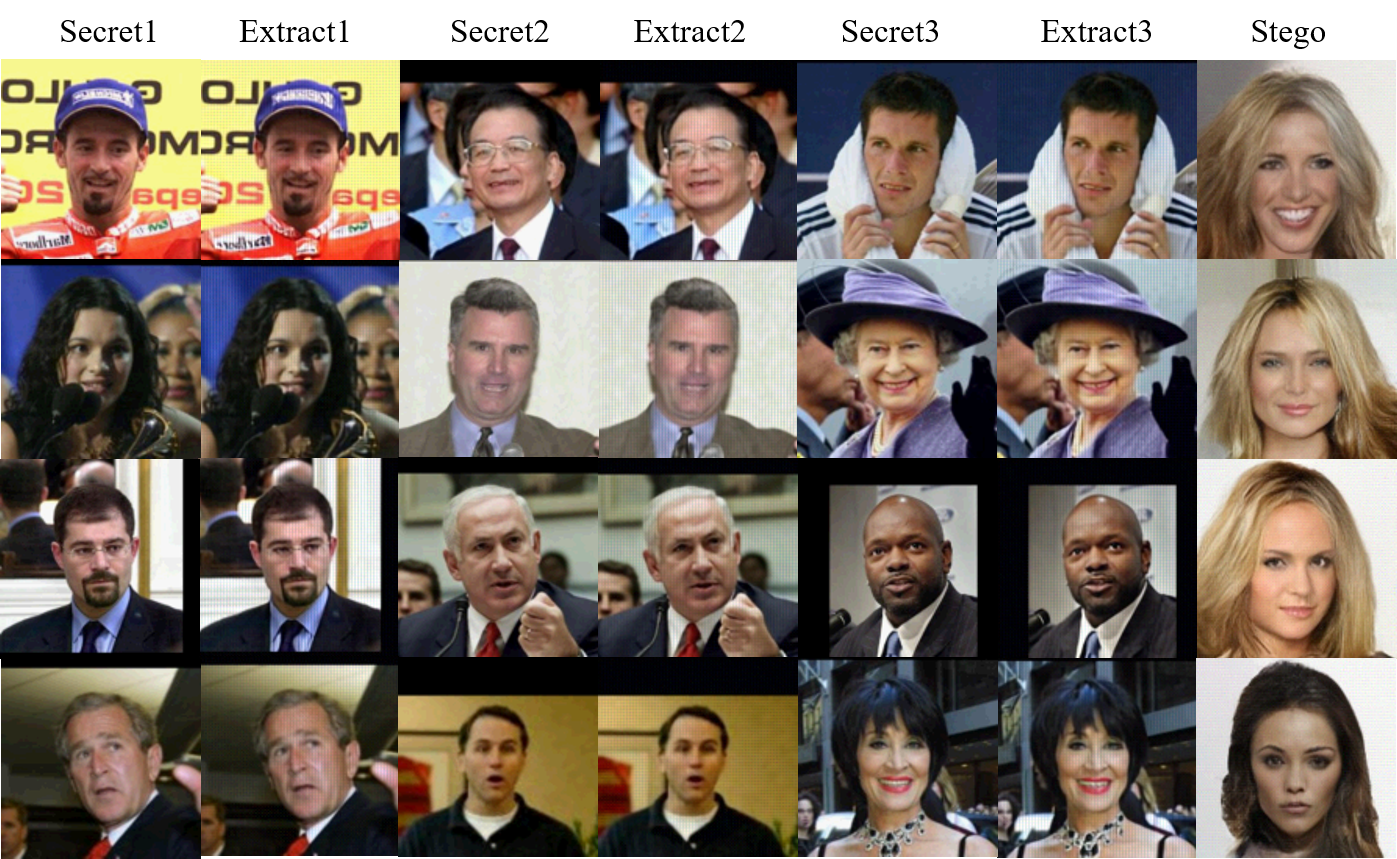}
    \caption{Multiple image hiding with $128 \times 128 \times 3$.}
    \label{fig6}
\end{figure*}

Most image hiding work can only hide a secret image to a cover image. However, it is not applicable to hide multiple secret images to an image, when specific integrated or sequentially related multiple images are not separable. Especially in image steganography without embedding, there is no work to do multiple image hiding and our method is proposed firstly to realize multi-image hiding without embedding.

In this section, we demonstrate the experimental results of DF-SWE for hiding multiple images in sizes of $ 64 \times 64 \times 3 $ and $ 128 \times 128 \times 3 $ in Figure \ref{fig5} and Figure \ref{fig6} respectively. $Secreti$ is the $i$-th secret image and $Extracti$ is the $i$-th extracted image from the Stego image (Stego) with respect to $Secreti$. 
It can be observed that, even there are three images (\textit{i.e.,} $BPP = 72$) hidden into the same stego image, the generated stego images remain natural. Moreover, the recovered secret images are nearly lossless.


\subsection{Domain generalization}
\renewcommand{\dblfloatpagefraction}{.9}
\begin{figure*}[htb]
\centering
    \includegraphics[width=.6\textwidth,height=.3\textwidth]{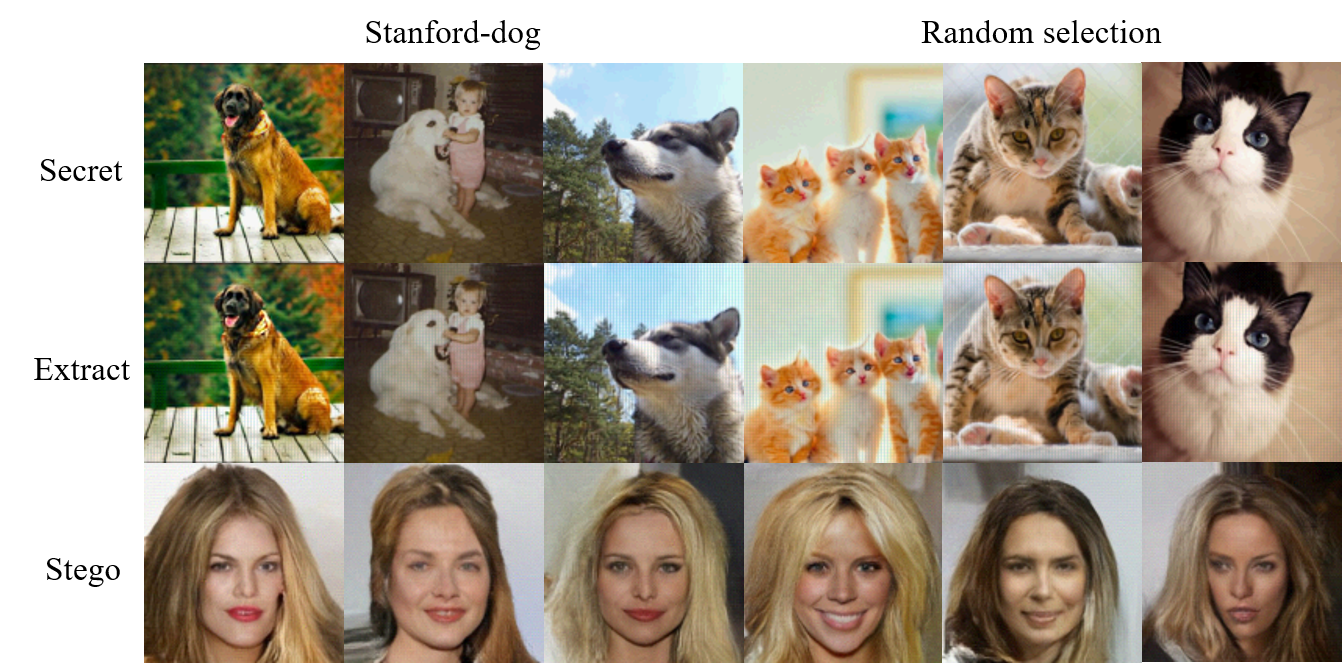}
    \caption{Data domain generalization.}
    \label{fig7}
\end{figure*}

Current image steganography usually requires that the secret images to be hidden are from the same domain of the samples used to train the steganography model.
However, it is expensive to train individual steganography models for images from new domains. 
Furthermore, collecting training data from particular domains could be difficult due to data privacy or other concerns.
Therefore, existing methods cannot achieve image steganography when accessing images from the same domain of the secret images is prohibited.
However, our method circumvents this limitation by its capability of domain generalization.
As shown in Figure \ref{fig7}, images in the first three columns on the left side are from the Stanford-dog dataset and the other images are randomly selected from the Internet. 
All the images have totally different distributions with that of images used to train DF-SWE.
According to Figure.~\ref{fig7}, DF-SWE can successfully hide and recover these images with a satisfactory visual quality. 
This property greatly boosts the capability of DF-SWE and makes it the first domain-agnostic steganography method.

\subsection{Ablation experiment}

\renewcommand{\dblfloatpagefraction}{.9}
\begin{figure*}[htb]
\centering
    \includegraphics[width=.9\textwidth,height=.55\textwidth]{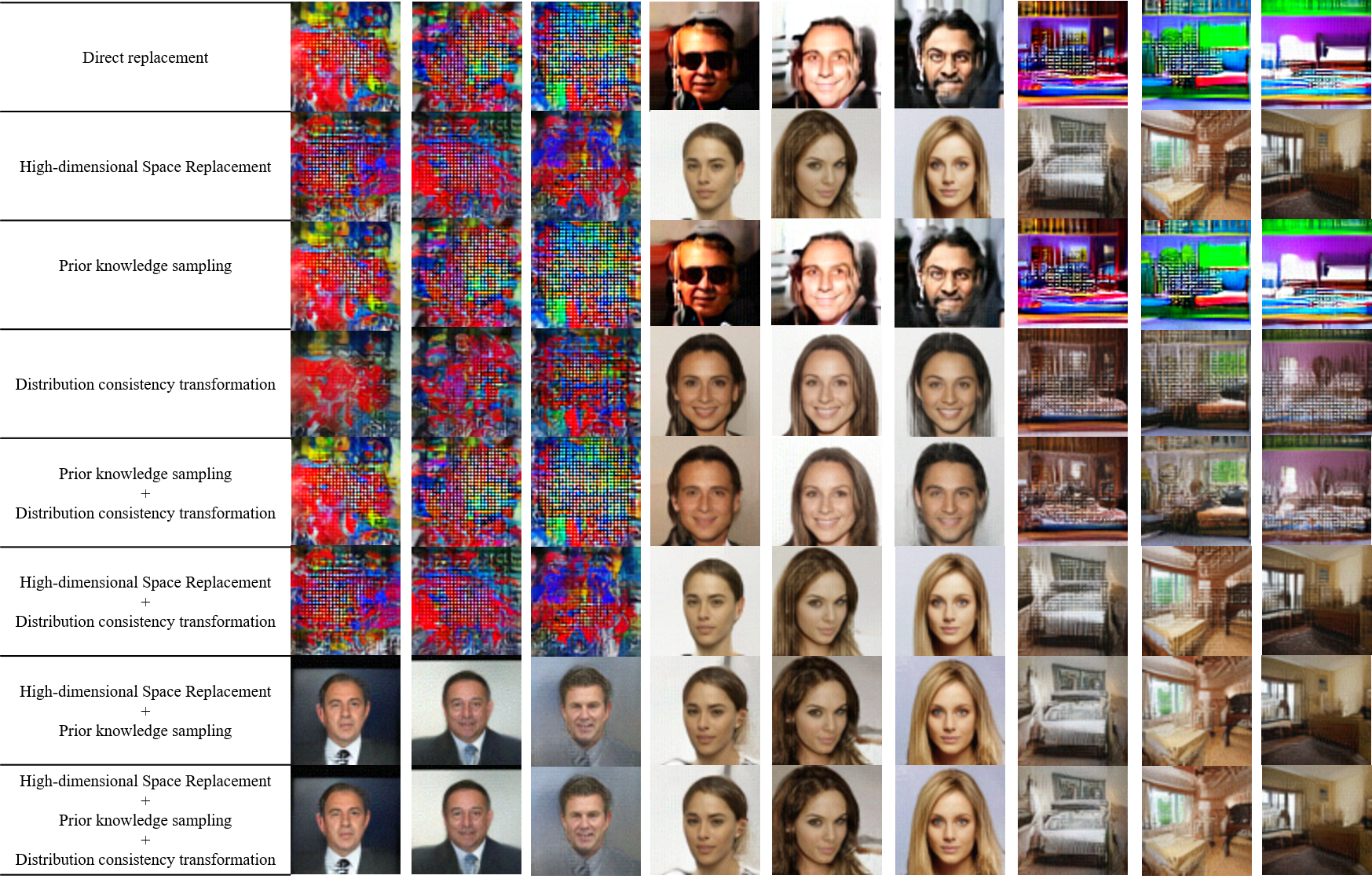}
    \caption{Ablation experiment of 3 tactics.}
    \label{fig4}
\end{figure*}

Figure \ref{fig4} performs an ablation analysis of 3 tactics employed by DF-SWE, which are prior knowledge sampling, high-dimensional space replacement and distribution consistency transformation. The first three, middle three and last three columns of images are the effect of different tactics on the LFW, CelebA and LSUN datasets, respectively. 

In the first row, the generated stego image is abnormal, particularly in the first three columns.
The main change of direct replacement tactic is that $\textsl{z}$ is replaced with $\textsl{z}_{se}$ directly without utilizing prior knowledge of $Model_{st}$. 
The high-dimensional space replacement is our proposed tactic shown in second row, which uses low-dimensional space of $\hat{\textsl{z}}^{se}$ to replace high-dimensional space of $\hat{\textsl{z}}^{st}$. We can see that high-dimensional space replacement effectively generates realistic images, but only this technique is not adequate from the abnormal images of the first three columns.
The prior knowledge sampling is our proposed method. In the third row, $\textsl{z}_{st}^{\prime}$ is replaced with $\textsl{z}_{se}$, which utilizes prior knowledge $Model_{st}$ and a multivariate
Gaussian latent-variable $\textsl{z}$.
The distribution consistency transformation is proposed to reduce the distortion from the difference between the two distributions. The fourth row is that $\textsl{z}$ is replaced with $\textsl{z}_{se}$, but $\textsl{z}_{se}$ is changed by the distribution consistency transformation. In the last three columns, the generated stego image is more normal than the first row.

The fifth row combines our proposed prior knowledge sampling and distribution consistency transformation. $\textsl{z}_{st}^{\prime}$ is replaced with $\textsl{z}_{se}$ which is modified by the distribution consistency transformation. In the last three columns, the quality of the generated image is a significant improvement compared with the first row.
In the sixth row, the first three columns indicate that only high-dimensional space replacement and distribution consistency transformation cannot generate a realistic image.
Compared with the second and seventh rows, the first three columns clearly show that prior knowledge sampling effectively improves the quality of the generated stego images.

In summary, the ablation experiments verify the effectiveness of our proposed method to circulate two latent flows and guarantee reversibility meanwhile. 

\section{Discussion and Future Work}
\label{Discussion and Future Work}

In this paper, we propose a novel double-flow-based steganography without embedding (DF-SWE) method for hiding large images. 
Specifically, we propose the reversible circulation of double flow to build a reversible bijective transformation between secret images and generated stego images. 
The reversible circulation ensures the small extraction error of the secret images and the high-quality of generated stego images.
Importantly, DF-SWE is the first SWE method that enables hiding images, or even multiple large images, into one stego image.
Specifically, the payload capacity of DF-SWE achieves $24-72 BPP$ and is $8000-16000$ times more than that of the other SWE methods.
In this way, DF-SWE provides a way to directly generate stego images without a cover image, which greatly improves the security of the secret images.
According to the experimental results, the proposed DF-SWE shows better hiding/recovering performance.
Intriguingly, DF-SWE can be generalized to hiding secret images from different domains with that of the training dataset.
This nice property indicates that DF-SWE can be deployed to privacy-critical scenarios in which the secret images are hidden from the provider of DF-SWE. Although our method achieves excellent performance for secret image recovery, the method is not completely lossless.

In the future, it is interesting to further explore the potential of SWE in lossless secret image recovery and multi-modal data hiding. The main challenge for multi-modal data hiding is how to map multi-modal data to the similarity multivariate Gaussian distribution. All these will be interesting future works to be explored.

\textbf{Acknowledgements: } 
This work is supported in part by the National Natural Science Foundation of China under Grant 62162067 and 62101480, Research and Application of Object detection based on Artificial Intelligence, in part by the Yunnan Province expert workstations under Grant 202205AF150145, in part by Yunnan Province Education Department Foundation under Grant No.2022j0008.



\bibliographystyle{IEEEtran}
\bibliography{IEEE}
\clearpage

\begin{IEEEbiography}[{\includegraphics[width=1in,height=1.25in,clip,keepaspectratio]{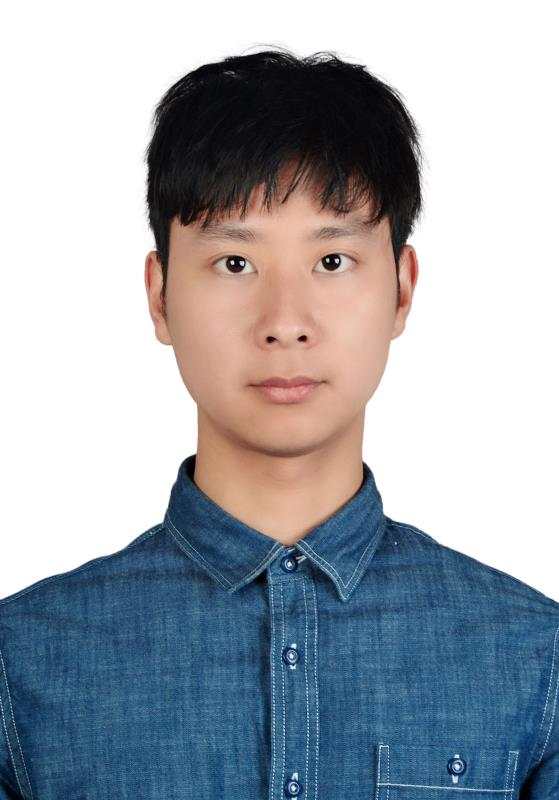}}]{Bingbing Song} is currently working toward the doctoral degree with the School of Information Science $ \& $ Engineering from Yunnan University. His current research interest includes developing deep learning algorithm for AI security and image steganography.
\end{IEEEbiography}

\begin{IEEEbiography}[{\includegraphics[width=1in,height=1.25in,clip,keepaspectratio]{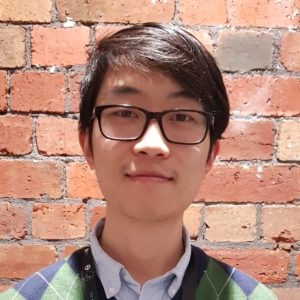}}]{Derui Wang} received his bachelor’s degree from Huazhong University of Science  $ \& $ Technology (HUST), China. He obtained his doctoral degree jointly from CSIRO’s Data61 and Swinburne University of Technology, Australia. He is currently a research scientist in CSIRO’s Data61. His primary research interest resides in the joint distribution of adversarial robustness verification, neural backdoors, and real-world security $ \& $ privacy issues of machine learning systems. He publishes papers in top journals and conferences, such as IEEE Transactions on Dependable and Secure Computing (TDSC), IEEE Transactions on Information Forensics and Security (TIFS), IEEE Symposium on Security and Privacy (S $ \& $ P), and The Network and Distributed System Security (NDSS) Symposium.
\end{IEEEbiography}

\begin{IEEEbiography}[{\includegraphics[width=1in,height=1.25in,clip,keepaspectratio]{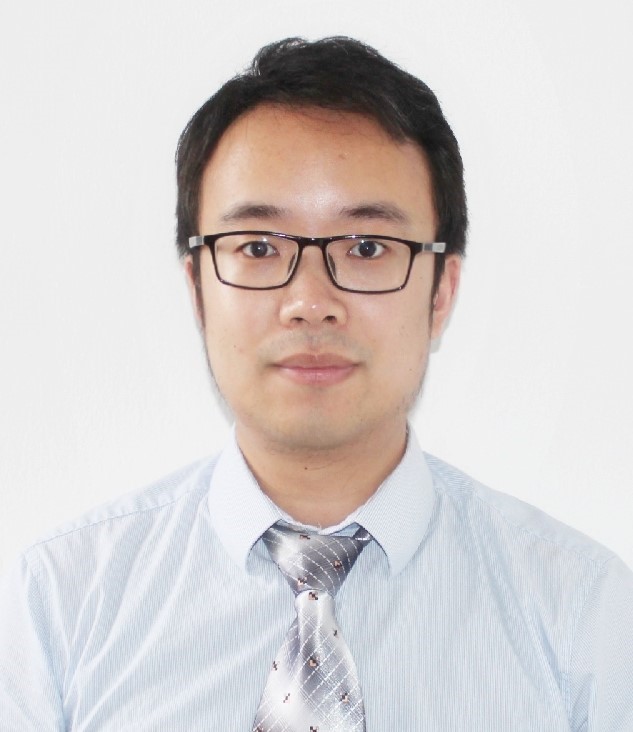}}]{Tianwei Zhang} is an assistant professor in School of Computer Science and Engineering, at Nanyang Technological University. His research focuses on computer system security. He is particularly interested in security threats and defenses in machine learning systems, autonomous systems, computer architecture and distributed systems. He received his Bachelor’s degree at Peking University in 2011, and the Ph.D degree in at Princeton University in 2017.
\end{IEEEbiography}

\begin{IEEEbiography}[{\includegraphics[width=1in,height=1.25in,clip,keepaspectratio]{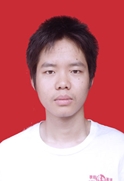}}]{Renyang Liu} received the BE degree in Computer Science from the Northwest normal University in 2017, He is a current Ph.D. candidate in School of Information Science and Engineering at Yunnan University, Kunming, China. His current research interest includes deep learning, adversarial attack and graph learning.
\end{IEEEbiography}

\begin{IEEEbiography}[{\includegraphics[width=1in,height=1.25in,clip,keepaspectratio]{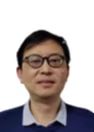}}]{Yu Lin} received the B.S. degree from Yunnan University, Kunming, China, and the Ph.D. degree in electron physics from China Academy of Electronic Sciences, Beijing, China. He is an Professor in Kunming Institute of Physics. His research interests include information fusion, infrared image processing, and object detection.
\end{IEEEbiography}

\begin{IEEEbiography}[{\includegraphics[width=1in,height=1.25in,clip,keepaspectratio]{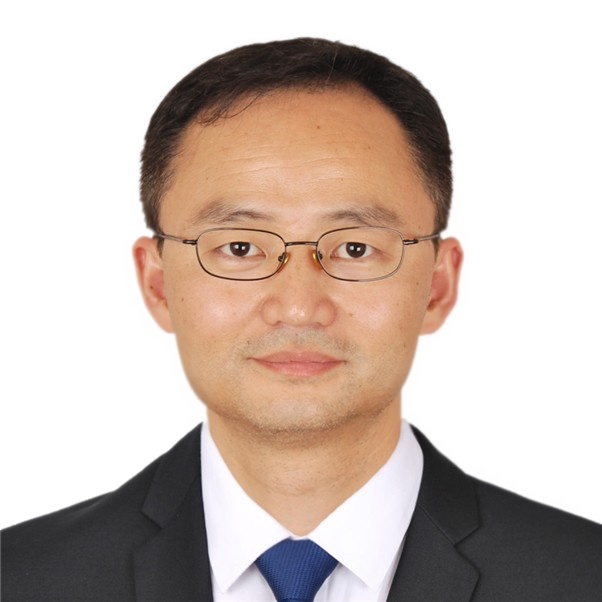}}]
{Wei Zhou} received his Ph.D. in computer science from the University of Chinese Academy of Science in 2008. Now he is a full professor at Yunnan University. He main research interests are about distributed computing, cloud computing and AI security.

\end{IEEEbiography}





\end{document}